\documentclass[sigconf]{acmart}
\usepackage{multirow}
\usepackage{graphicx}
\usepackage{caption}
\usepackage{subcaption}
\usepackage{amsmath}
\usepackage{amsthm}
\usepackage{booktabs}
\usepackage{algpseudocode}
\usepackage{xcolor}
\usepackage{algorithm}
\usepackage{pifont}
\usepackage{color}
\usepackage{tcolorbox}
\usepackage{enumitem}
\usepackage{balance}

\usepackage{changes}
\definechangesauthor[name={feng}, color=orange]{feng}

\AtBeginDocument{%
  \providecommand\BibTeX{{%
    \normalfont B\kern-0.5em{\scshape i\kern-0.25em b}\kern-0.8em\TeX}}}

\copyrightyear{2021}
\acmYear{2021}
\setcopyright{acmcopyright}\acmConference[MM '21]{Proceedings of the 29th ACM International Conference on Multimedia}{October 20--24, 2021}{Virtual Event, China}
\acmBooktitle{Proceedings of the 29th ACM International Conference on Multimedia (MM '21), October 20--24, 2021, Virtual Event, China}
\acmPrice{15.00}
\acmDOI{10.1145/3474085.3475662}
\acmISBN{978-1-4503-8651-7/21/10}

\begin{document}
\fancyhead{}

\title{Similar Scenes arouse Similar Emotions: Parallel Data Augmentation for Stylized Image Captioning}


\author{Guodun Li,\quad Yuchen Zhai,\quad Zehao Lin, \quad Yin Zhang$^{*}$}

\makeatletter
\def\authornotetext#1{
\if@ACM@anonymous\else
    \g@addto@macro\@authornotes{
    \stepcounter{footnote}\footnotetext{#1}}
\fi}
\makeatother
\authornotetext{Corresponding author: Yin Zhang}

\affiliation{
 \institution{College of Computer Science and Technology, Zhejiang University \city{Hangzhou} \country{China}}
 }
\email{{guodun.li, zhaiyuchen, georgelin, zhangyin98}@zju.edu.cn}

\def\authors{Guodun Li, Yuchen Zhai, Zehao Lin, Yin Zhang}

\renewcommand{\shortauthors}{Guodun Li, et al.}


\begin{abstract}
Stylized image captioning systems aim to generate a caption not only semantically related to a given image but also consistent with a given style description. One of the biggest challenges with this task is the lack of sufficient paired stylized data. Many studies focus on unsupervised approaches, without considering from the perspective of data augmentation. We begin with the observation that people may recall similar emotions when they are in similar scenes, and often express similar emotions with similar style phrases, which underpins our data augmentation idea. In this paper, we propose a novel Extract-Retrieve-Generate data augmentation framework to extract style phrases from small-scale stylized sentences and graft them to large-scale factual captions. First, we design the emotional signal extractor to extract style phrases from small-scale stylized sentences. Second, we construct the plugable multi-modal scene retriever to retrieve scenes represented with pairs of an image and its stylized caption, which are similar to the query image or caption in the large-scale factual data. In the end, based on the style phrases of similar scenes and the factual description of the current scene, we build the emotion-aware caption generator to generate fluent and diversified stylized captions for the current scene. Extensive experimental results show that our framework can alleviate the data scarcity problem effectively. It also significantly boosts the performance of several existing image captioning models in both supervised and unsupervised settings, which outperforms the state-of-the-art stylized image captioning methods in terms of both sentence relevance and stylishness by a substantial margin.
\end{abstract}



\maketitle

\begin{figure}
    \centering
    \includegraphics[width=0.8\linewidth]{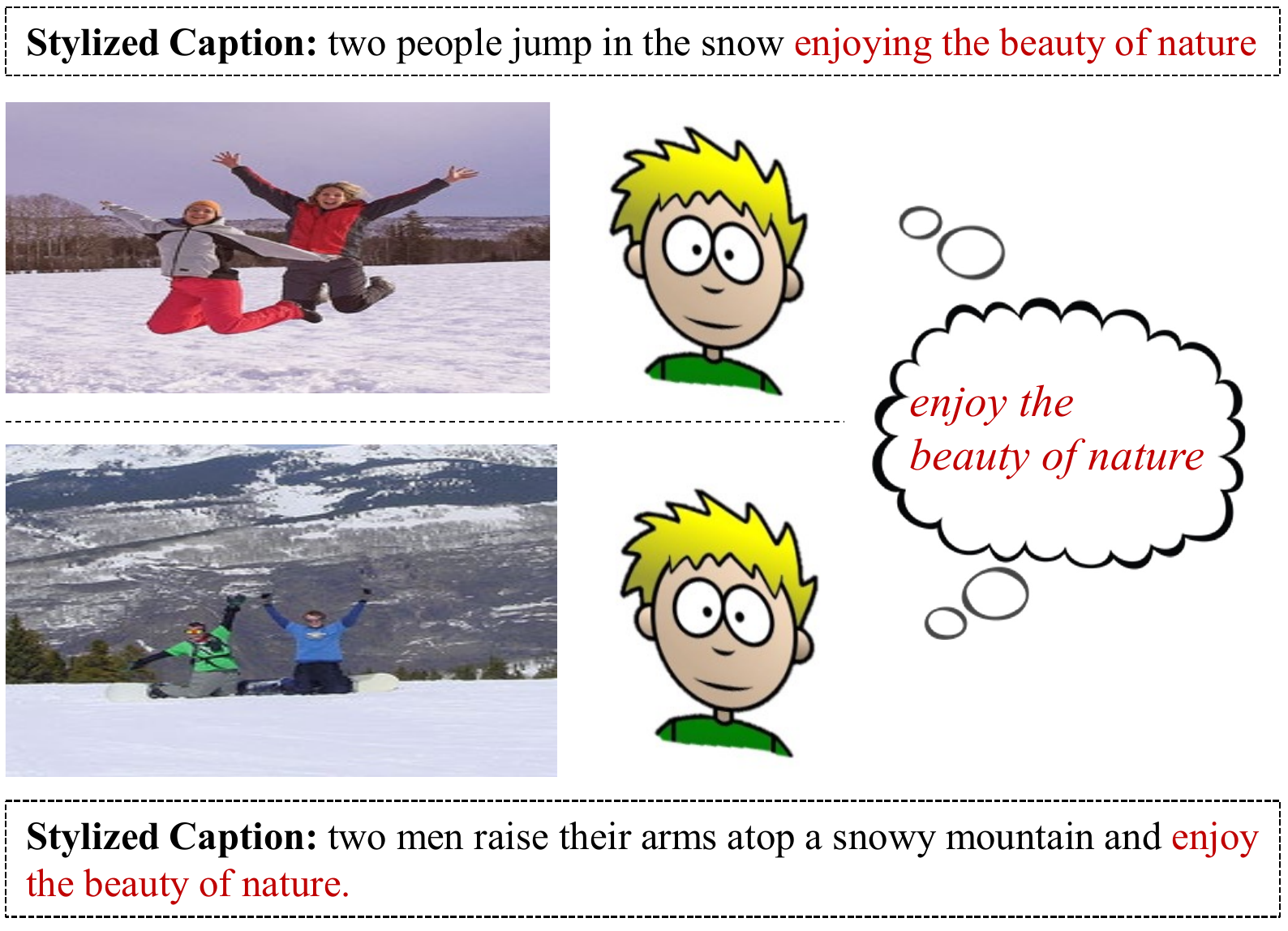}
    \caption{An example that when people are facing similar scenes, they arouse similar emotions, and express similar emotions with similar style phrases. A group of words in red is the style phrase of the stylized caption.}
    \label{fig:introduction}
\end{figure}

\section{Introduction}
Stylized image captioning~\cite{gan2017stylenet,mathews2016senticap} aims to generate a natural language description that is semantically related to a given image and consistent with a given style description. Since most image captioning models are data-hungry, one of the biggest challenges with this task is the lack of a large-scale parallel dataset containing dozens of pairs of an image and its stylized caption. It is time consuming and expensive to manually compose stylized captions for large-scale images. Hence, previous studies focus on unsupervised approaches, relying on unpaired stylized texts and existing large-scale factual image-caption dataset.

Limitations of such unsupervised approaches are as follows: 1) Most of the existing works~\cite{mathews2018semstyle, chen2019unsupervised} learn the latent style vector based on unpaired stylized texts. However, in real-world applications, the style domain of large available unpaired texts are often far away from target style domain, making the trained models not able to perform well. 2) These well-designed unsupervised approaches are hard to employ state-of-the-art multi-modal pre-trained models~\cite{zhang2020vinvl,li2020oscar}, which seriously limits the performance of this task. 3) It is hard to balance the use of the factual information in the large-scale image-caption corpus and style knowledge acquired from small-scale style texts because of the imbalanced data. 

Contrasted with previous methods, we investigate how to improve the performance of this task from a different and orthogonal angle, i.e. augmenting more high-quality paired stylized data to alleviate the data scarcity problem. Our work begins with the real experience during social interactions, where people arouse similar emotions when they are facing similar scenes, which is called the ``context effect''~\cite{godden1975context, dulsky1935effect, abernethy1940effect} in psychology. This is because a similar visual context makes people extract similar emotions from past memory~\cite{godden1975context}. Further, people often express similar emotions with similar stylized phrases. Figure 1 shows two image-caption pairs, those images are similar to each other. And both captions share the same style phrase \textit{``enjoy the beauty of nature''} to express the same emotion.

Based on the ``context effect'', we propose a novel \emph{Extract-Retrieve-Generate} data augmentation framework (ERG) to yield more paired stylized data in an automated manner. ERG is composed of three modules: 1) \emph{Style phrase extractor} discerns linguistic style phrases from existing stylized captions in the small parallel stylized caption dataset. For example, for an input stylized caption \textit{``two men raise their arms atop a snowy mountain and enjoy the beauty of nature.''}, the module is able to extract the style phrase \textit{``enjoy the beauty of nature''}. 2) \emph{Multi-modal scene retriever} returns a list of similar scenes in the small stylized data by leveraging the state-of-the-art multi-modal transformer-based models to compute the image-to-image, image-to-text, text-to-image, text-to-text similarities between sample pairs in the large factual corpus and those in the small stylized corpus, inspired by the observation that people often use similar images, texts, and other perceptual inputs as clues to recall past similar scenes. 3) \emph{Emotion-aware caption generator} produces stylized image captions, given images in the large-scale factual dataset and the style phrases of scenes retrieved in the small stylized corpus.

The main contributions of this work are summarized as follows:
\begin{itemize}
    \item To the best of our knowledge, this work is the first effort towards data augmentation of stylized image captioning. Our approach can be combined with any existing supervised, unsupervised and transformer-based approaches.
    \item We propose a novel Extract-Retrieve-Generate data augmentation framework to automatically generate high-quality and diversified stylized captions for images in the large-scale factual dataset by leveraging the visually or textually relevant scenes and their style phrases in the stylized corpus.
    \item Comprehensive experimental results on two widely-used stylized image-caption datasets show that our framework can alleviate the data scarcity problem significantly. It also effectively boosts the performance of several existing image captioning models in both supervised and unsupervised settings, which outperforms the state-of-the-art stylized image caption methods by a substantial margin.
\end{itemize}

\section{Related Work}
\label{sec:related}
\subsection{Stylized Image Captioning}
Image captioning is a hot and fundamental topic~\cite{hossain2019comprehensive,sheng2019generating,10.1145/3343031.3350996,10.1145/3343031.3350943} of intersection of advances in computer vision and natural language processing. Recently, novel image caption models~\cite{vinyals2015show,Yao2019AttentionAwarePS,Yin2019ContextAA} can generate captions from both visual latent space or multi-modal (images and texts) latent space by analyzing the visual content of the image first, e.g. using pre-trained image model like ResNet~\cite{szegedy2017inception}, and then generate image captions from the visual content using a language model~\cite{xu2015show,yao2017boosting,you2016image}.

Stylized image captioning~\cite{mathews2016senticap} takes a further step, aiming at generating captions in target styles. Most existing work~\cite{gan2017stylenet, mathews2018semstyle, you2018image, chen2019unsupervised, guo2019mscap, Zhao2020MemCapMS} use non-parallel stylized image-caption data and the others~\cite{mathews2016senticap,chen2018factual, shuster2019engaging} use parallel stylized corpus. StyleNet~\cite{gan2017stylenet} proposes to generate humorous and romantic captions by factorizing the
input weight matrices to contain a style-specific factor matrix. This work mainly focuses on capturing multiple linguistic styles information. Similarly, SF-LSTM~\cite{chen2018factual} proposes to learn two groups of matrices to capture the factual and stylized knowledge, respectively. You et al.\cite{you2018image} choose another way to inject
sentiments into image captions and control the sentiment by providing different sentiment labels in semi-supervised mode without requirement of parallel stylized data. SemStyle~\cite{mathews2018semstyle} develops a model that learns to generate visually relevant styled captions from a large corpus of stylized text without aligned images. Nezami et al.\cite{nezami2019towards} propose an image captioning
model called ATTEND-GAN which has an
attention-based caption generator and adversarial training mechanism to generate correlated captions as well as more human-like variability of stylistic patterns. Similarly, Guo et al.~\cite{guo2019mscap} propose an adversarial learning framework with several auxiliary modules on unpaired stylized texts to generate captions in multiple styles. Recent advances in this task is MemCap\cite{Zhao2020MemCapMS}, which explicitly encodes the knowledge about linguistic styles learned from unpaired stylized corpus with memory mechanism. Contrasted with previous works, we investigate how to improve the existing models from a different angle, i,e. producing more parallel stylized data with data augmentation techniques.

\subsection{Data Augmentation}
Data Augmentation is an effective way to improve the performance of neural models,  which encompasses a set of techniques to increase the size and improve the quality of datasets. Such techniques reduce the cost of dataset collection, and usually boost the model performance on desired tasks. It has been widely applied in computer vision, natural language processing, and multimedia areas, especially for tasks without sufficient data~\cite{li2020textaug, luo2020imageaug, zhang2020videoaug}. In image captioning, data augmentation has been used in the previous work~\cite{wang2018captionaug, Katiyar2020captionaug}, which focus on the image transformations to avoid over-fitting. 

However, for stylized image captioning, due to the lack of parallel data, many studies focus on unsupervised approaches, which seriously limits the performance of powerful models such as pre-trained transformers~\cite{li2020oscar, guo2020san, zhang2020vinvl}. There is no related work concerning data augmentation. Existing oversimple augmentation techniques based on a single modal such as image transformations can not contribute to the linguistic style of captions. To solve this pain point, we propose the Extract-Retrieve-Generate framework for paired stylized data augmentation in an automatic manner.

\begin{figure*}
    \centering
    \includegraphics[width=0.95\linewidth]{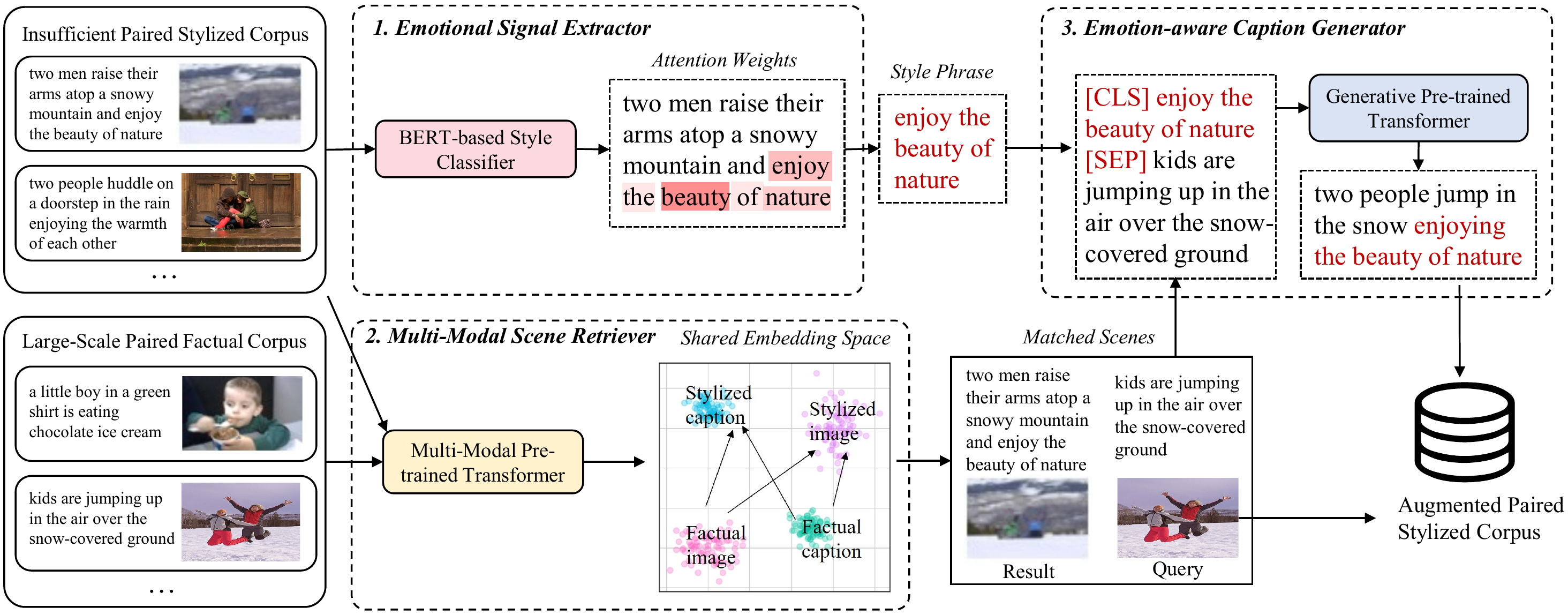}
    \caption{Overview of the proposed Extract-Retrieve-Generate data augmentation framework. First, the Emotional Signal Extractor extracts style phrases of the stylized sentences according to the attention weights. Second, for each factual image-caption pair, the Multi-Modal Scene Retriever retrieves the similar stylized image-caption pair in a shared embedding space through four types of retrieval method, i.e., image-to-image, image-to-text, text-to-image, and text-to-text retrieval. Finally, given the style phrase of the retrieved image-caption pair and the factual caption of the queried pair, the Emotion-aware Caption Generator generates a stylized caption for the image of the queried pair.}
    \label{fig:overview}
\end{figure*}

\section{Approach}
\subsection{Overview}
A stylized image captioning model is expected to generate a caption not only semantically related to the given image but also consistent with a given linguistic style. Suppose we have an insufficient training set $\mathcal{D}_s=\{(x^s_i, y_i^s)|_i\}$ of paired stylized data, where $x^s_i$ and $y_i^s$ denote the $i$-th image and its corresponding stylized caption, respectively, and a large-scale paired factual corpus $\mathcal{P}=\{(x^f_j,y_j^f)|_j\}$,  where $x^f_j$ and $y_j^f$ denote the $j$-th image and its corresponding factual caption. The objective of our augmentation task is to produce large-scale paired stylized data $\mathcal{D}_a$, based on large-scale factual corpus $\mathcal{P}$ and insufficient paired stylized data $\mathcal{D}_s$. With original paired stylized data $\mathcal{D}_s$ and augmented data $\mathcal{D}_a$, we can train a better stylized image captioning model in a supervised way.

Figure~\ref{fig:overview} shows the whole pipeline of the proposed augmentation method. It consists of three components: 1) \emph{The emotional signal extractor}, which aims to extract the style phrase of $y^s$. 2) \emph{The multi-modal scene retriever}, which aims to retrieve a set of similar scenes in the small stylized dataset, given an image in the large factual dataset. 3) \emph{The emotion-aware caption generator}, which leverages the style phrases of similar scenes retrieved above to generate stylized captions for images in the large-scale factual corpus $\mathcal{P}$. With the above components, we are able to form a new large-scale paired stylized data  $\mathcal{D}_a=\{(x_k^f,\hat{y}_k^s)|_k\}$ to augment the original insufficient training set $\mathcal{D}_s$. In the following sections, we will describe these three components in detail.

\label{sec:approach}

\subsection{Emotional Signal Extractor}
\label{sec:ese}
Given an image $x^s$ and its corresponding stylized caption $y^s$, the module is responsible for extracting the style phrase of the caption in an unsupervised way.

The structure of our extractor is inspired by~\cite{li2020attack} and~\cite{Sudhakar2019delete}, which recognize vulnerable words for text adversarial attack and attribute words for text style transfer by iteratively deleting words and observing the changes of output. Similarly, the style phrase of a stylized caption is often localized to certain words or phrases. A style classifier will be confused about its style if we remove these words, leading to significantly changes of output.

With the above observation, our extractor is modeled as a well-trained style classifier, determining which word has a high impact on the final output. Formally, let $y^s=[w_0,...,w_t,...]$ denote the input stylized caption with style label $s$, and $I(w_t)$ is defined as the importance score of $w_t$ to measure how much it contributes to the style of the caption. We rank all the words according to their importance scores in descending order and only extract $\epsilon$ proportion of the most important words as the style phrase. $\epsilon$ is also referred to as extraction proportion.

In practice, we adopt a BERT-based classifier~\cite{Devlin2019bert} fine-tuned on the stylized captions as our extractor because of its strong performance on various NLP downstream tasks. The importance score is defined as the attention weights of other input tokens that correspond to $[CLS]$ token. Since BERT is stacked with multiple layers and multiple heads of transformer blocks, we can obtain multiple attention weights of words corresponding to every layer $l$ and head $h$, formally: 
\begin{equation}
    I_{h,l}(w_t) =  \mathrm{softmax}_{w_t\in y^s}(\frac{Q_{h,l,[CLS]}K_{h,l,w_t}^T}{\sqrt{d_k}}),
\end{equation}
where $Q_{h,l,[CLS]}$ and $K_{h,l,w_t}$ are the query vector of $[CLS]$ and the key vector of $w_t$ on the layer $l$ and head $h$, and $d_k$ is a scaling factor.

However, every layer and head encode different aspects of semantic and linguistic structure, we need to determine which attention weights are really related to the linguistic style aspect of captions. Concretely, for each layer and head, we remove $\epsilon$ percent of the most important words denoted as $w_{\epsilon}$ to form a new reduced caption $y^s_{/w_{\epsilon}}$. Following~\cite{Sudhakar2019delete},  we define the confidence score $c(h,l,y^s)$  for each layer and head to measure the changes of output probability, formally:
\begin{equation}
c(h,l,y^s) = \frac{p(s|y^s_{/w_{\epsilon}})}{\sum_{s'\in S'}{p(s'|y^s_{/w_{\epsilon}})}},    
\end{equation}
where $S'=S-\{s\}$ is the remaining style labels of the label set $S$. The lower the confidence score, the more confused the classifier is about its style, and the greater the change of output. Finally, we choose the layer $l_s$ and head $h_s$ that minimize the average confidence score of all stylized captions in $\mathcal{D}_s$ to form the importance score $I_{h_s,l_s}(w_t)$.

For each stylized caption, measured by $I_{h_s,l_s}(w_t)$, we extract $\epsilon$ percent of the most important words in the caption $y^s$ as the style phrase $e^s$. In this way, we update the paired stylized data $\mathcal{D}_s$ to $\mathcal{D'}_s=\{(x_i^s, y_i^s, e_i^s)|_i\}$.

\subsection{Multi-Modal Scene Retriever}
\label{sec:mcr}

To follow the intuition that similar scenes may be described in a similar stylized expression, the module is responsible for retrieving a set of style phrases from $\mathcal{D}'_s$ for images in the corpus $\mathcal{P}$ by utilizing multi-modal cues.

\begin{table}[]
\centering
\caption{Four types of scene retrieval, i.e., image-to-image, image-to-text, text-to-image, and text-to-text retrieval.}
\resizebox{\linewidth}{!}{%
\begin{tabular}{lll}
\hline
Retrieval Type       & Query       & Target                                       \\ \hline
image-to-image (i2i) & image $x^f$ & image $x^s$ with style phrase $e^s$ \\
text-to-text (t2t)   & text $y^f$  & text $y^s$ with style phrase $e^s$  \\
image-to-text (i2t)  & image $x^f$ & text $y^s$ with style phrase $e^s$ \\
text-to-image (t2i)  & text $y^f$  & image $x^s$ with style phrase $e^s$  \\ \hline
\end{tabular}%
}
\label{tab:retrieval-type}
\end{table}

As shown in Figure~\ref{fig:overview}, we adopt a multi-modal pre-trained transformer called OSCAR~\cite{li2020oscar} as a unified multi-modal encoder $U(\cdot)$ to map all the texts and images into $d$-dimensional vectors in a shared embedding space. 

Specifically, for each input image, we firstly extract $K$ region-level visual features, with the Faster R-CNN~\cite{Ren2015rcnn} model pre-trained on Visual Genomes~\cite{Krishna2017genome} as the input of pre-trained transformer. Each image and caption are passed separately through the pre-trained transformer model, which outputs the final hidden representation of $[CLS]$ token as the respective image representation $E_x$ and caption representation $E_y$. The similarity is defined as the cosine of the angle between their embedding vectors, formally:
\begin{equation}
\mathrm{sim}(x,y) = \frac{E_{x}\cdot E_y}{\|E_x\|\|E_y\|}
\end{equation}

Subsequently, we adopt an efficient similarity search tool called Faiss~\cite{JDH17} to implement the retrieval process through four types of scene retrieval, as shown in Table~\ref{tab:retrieval-type}. 

In this way, we can retrieve more diverse style phrases for images in large-scale factual data $\mathcal{P}$.

\subsection{Emotion-aware Caption Generator}
With the retrieved style phrases, the caption generator module is responsible for generating stylized captions for images in the corpus $\mathcal{P}$ to construct new paired stylized samples.

To achieve this, we regard the factual caption $y^f$ in the corpus $\mathcal{P}$ as the content of corresponding images and the retrieved style phrase $e^s$ as the potential emotional prompt when describing the image. The conditional probability of generating captions given above information is defined as follows:
\begin{equation}
    p(\hat{y}^s|x^f)\approx p(\hat{y}^s|y^f,e^s)
\end{equation}

Motivated by the recent remarkable performance of GPT~\cite{radford2019gpt}, a powerful generative transformer pre-trained on large-scale unlabeled texts, we adopt GPT-2 as our generator and fine-tune it to learn the above conditional probability.

Let $\theta$ denote the parameters of GPT-2. In the fine-tuning process, for each stylized caption $y^s$ in the $\mathcal{D}_s$, we first extract the style phrase $e^s$ and regard the rest part $y^s_{/w_\epsilon}$ as the content of the image. Thus, the fine-tuning objective is to reconstruct the original stylized caption, i.e., minimizing the following objective:
\begin{equation}
   L(\theta)=-\sum_{t=1}^T\log p(y^s_t|y^s_{1:t-1};e^s;y^s_{/w_\epsilon}) 
\end{equation}
When inferring a new stylized caption, a sample of the process is illustrated as follows:
\begin{enumerate}[leftmargin=*]
    \item Retrieved style phrase: \textit{enjoy the beauty of nature}
    \item Factual caption in the corpus $\mathcal{P}$: \textit{Kids are jumping up in the air over the snow-covered ground}
    \item Input representation: $[CLS]$ enjoy the beauty of nature $[SEP]$ Kids are jumping up in the air over the snow-covered ground
    \item Output: two people jump in the snow enjoying the beauty of nature
\end{enumerate}

Besides, we adopt the following three criteria to guarantee the quality of generated data: 1) the generated caption sentence can be classified by the BERT style classifier (see Sect. 4.4) as that target style, 2) its perplexity < 80 (see Sect. 4.4), 3) scene retrieval similarity score > 0.6 (see Fig.3). Those valid generated caption samples are used to construct paired augmented data.

\begin{table*}[]
\centering
\caption{Performance comparisons on the test splits of four styles. ``Original'' and ``Augmentation'' mean the models use the original or augmented training data. B-n, M, C, cls., and ppl. are short for BLEU-n, METEOR, CIDEr, style classification
accuracy, and perplexity, respectively. $\Delta$C and $\Delta$cls. measure the performance gains in terms of CIDEr and cls. metrics.}
\begin{tabular}{c|l|llllll|llllllll}
\hline
\multicolumn{1}{l|}{\multirow{2}{*}{Style}} & \multirow{2}{*}{Model} & \multicolumn{6}{c|}{Original}                                                                                                       & \multicolumn{8}{c}{Augmentation}                                                                                                                                                                               \\ \cline{3-16} 
\multicolumn{1}{l|}{}                       &                        & \multicolumn{1}{c}{B-1} & \multicolumn{1}{c}{B-3} & \multicolumn{1}{c}{M} & \multicolumn{1}{c}{C} & \multicolumn{1}{c}{cls.} & ppl.  & \multicolumn{1}{c}{B-1} & \multicolumn{1}{c}{B-3} & \multicolumn{1}{c}{M} & \multicolumn{1}{c}{C} & \multicolumn{1}{c}{cls.} & \multicolumn{1}{l|}{ppl.}  & $\Delta$C(\%)                & $\Delta$cls.(\%)             \\ \hline
\multirow{5}{*}{Humor}                      & NIC                    & 27.3                    & 8.3                     & 10.9                  & 39.2                  & 82.8                     & 6.0  & 28.4                    & 9.6                     & 11.8                  & 45.2                  & 97.5                     & \multicolumn{1}{l|}{9.2}  & \textbf{+6.0}   & \textbf{+14.7}  \\
                                            & Up-Down                & 29.9                    & 9.9                     & 12.2                  & 50.5                  & 80.7                     & 8.6  & 30.3                    & 10.7                    & 13.3                  & 54.2                  & 98.6                     & \multicolumn{1}{l|}{8.7}  & \textbf{+3.7}   & \textbf{+17.9}  \\
                                            & StyleNet               & 24.6                    & 7.3                     & 10.3                  & 34.7                  & 42.5                     & 7.3  & 24.6                    & 6.8                     & 9.8                   & 28.3                  & 96.2                     & \multicolumn{1}{l|}{7.6}  & -6.4  & \textbf{+53.7}  \\
                                            & SAN                    & 27.6                    & 8.1                     & 11.2                  & 38.7                  & 87.8                     & 8.4  & 29.5                    & 9.9                     & 12.5                  & 47.2                  & 99.4                     & \multicolumn{1}{l|}{13.7} & \textbf{+8.5}   & \textbf{+11.6}  \\
                                            & VinVL                  & 26.5                    & 7.6                     & 11.3                  & 38.2                  & 89.1                     & 12.9 & 28.9                    & 9.0                     & 12.2                  & 44.4                  & 99.7                     & \multicolumn{1}{l|}{12.2} & \textbf{+6.2}   & \textbf{+10.6}  \\ \hline
\multirow{5}{*}{Roman}                      & NIC                    & 28.1                    & 8.7                     & 11.2                  & 41.6                  & 86.5                     & 5.4  & 31.0                    & 9.9                     & 12.4                  & 47.2                  & 97.5                     & \multicolumn{1}{l|}{8.0}  & \textbf{+5.6}   & \textbf{+11.0}  \\
                                            & Up-Down                & 30.7                    & 10.7                    & 12.9                  & 55.5                  & 86.4                     & 7.7  & 32.8                    & 11.5                    & 13.6                  & 57.9                  & 97.4                     & \multicolumn{1}{l|}{8.6}  & \textbf{+2.4}   & \textbf{+11.0}  \\
                                            & StyleNet               & 24.8                    & 7.4                     & 10.3                  & 34.4                  & 57.1                     & 6.9  & 25.8                    & 7.0                     & 10.2                  & 30.3                  & 92.7                     & \multicolumn{1}{l|}{5.9}  & -4.1  & \textbf{+35.6}  \\
                                            & SAN                    & 28.3                    & 8.7                     & 11.5                  & 42.1                  & 90.9                     & 9.1  & 30.9                    & 10.9                    & 13.0                  & 53.3                  & 99.6                     & \multicolumn{1}{l|}{13.1} & \textbf{+11.2}  & \textbf{+8.7}   \\
                                            & VinVL                  & 28.7                    & 8.3                     & 11.8                  & 41.2                  & 96.6                     & 10.1 & 31.1                    & 10.1                    & 13.1                  & 52.1                  & 96.8                     & \multicolumn{1}{l|}{11.0} & \textbf{+10.9}  & \textbf{+0.20}  \\ \hline
\multirow{5}{*}{Pos}                        & NIC                    & 40.3                    & 15.2                    & 11.8                  & 29.9                  & 99.4                     & 5.2  & 50.5                    & 22.6                    & 17.3                  & 69.1                  & 100.0                    & \multicolumn{1}{l|}{12.2} & \textbf{+39.2}  & \textbf{+6.0}   \\
                                            & Up-Down                & 49.6                    & 20.7                    & 15.9                  & 52.3                  & 92.8                     & 9.8  & 52.4                    & 24.4                    & 18.1                  & 77.7                  & 100.0                    & \multicolumn{1}{l|}{10.3} & \textbf{+25.4}  & \textbf{+7.2}   \\
                                            & StyleNet               & 43.9                    & 13.0                    & 11.6                  & 30.8                  & 0.0                      & 10.3 & 33.2                    & 8.7                     & 8.3                   & 14.9                  & 100.0                    & \multicolumn{1}{l|}{4.9}  & -15.9 & \textbf{+100.0} \\
                                            & SAN                    & 44.0                    & 17.2                    & 13.5                  & 41.5                  & 93.2                     & 6.4  & 53.0                    & 23.4                    & 18.1                  & 72.0                  & 100.0                    & \multicolumn{1}{l|}{11.7} & \textbf{+30.5}  & \textbf{+6.8}   \\
                                            & VinVL                  & 45.6                    & 15.6                    & 14.4                  & 43.3                  & 76.9                     & 12.6 & 53.4                    & 23.4                    & 18.0                  & 75.9                  & 100.0                    & \multicolumn{1}{l|}{12.4} & \textbf{+32.6}  & \textbf{+23.1}  \\ \hline
\multirow{5}{*}{Neg}                        & NIC                    & 44.5                    & 15.7                    & 14.2                  & 33.2                  & 96.0                     & 5.8  & 51.4                    & 19.9                    & 17.4                  & 62.2                  & 100.0                    & \multicolumn{1}{l|}{12.5} & \textbf{+29.0}  & \textbf{+4.0}   \\
                                            & Up-Down                & 51.0                    & 19.6                    & 16.5                  & 50.8                  & 91.7                     & 11.3 & 52.6                    & 21.6                    & 18.0                  & 68.3                  & 100.0                    & \multicolumn{1}{l|}{8.5}  & \textbf{+17.5}  & \textbf{+8.3}   \\
                                            & StyleNet               & 43.5                    & 12.9                    & 11.7                  & 27.8                  & 10.1                     & 8.5  & 36.3                    & 8.2                     & 10.6                  & 13.6                  & 100.0                    & \multicolumn{1}{l|}{9.0}  & -14.2 & \textbf{+89.9}  \\
                                            & SAN                    & 44.6                    & 15.2                    & 14.0                  & 36.0                  & 92.3                     & 8.0  & 51.2                    & 20.5                    & 17.6                  & 67.0                  & 100.0                    & \multicolumn{1}{l|}{14.8} & \textbf{+31.0}  & \textbf{+7.7}   \\
                                            & VinVL                  & 44.6                    & 14.6                    & 15.5                  & 41.4                  & 88.1                     & 15.1 & 53.1                    & 21.2                    & 18.6                  & 70.0                  & 100.0                    & \multicolumn{1}{l|}{13.7} & \textbf{+28.6}  & \textbf{+11.9}  \\ \hline
\end{tabular}
\label{tab:res-main}
\end{table*}

\section{Experiments}
\label{sec:exp}
\subsection{Experimental Datasets}
To validate and analyze the effectiveness of our data augmentation framework on the stylized image caption task, we conduct our experiments on two popular stylized image caption datasets: FlickrStyle10K and SentiCap. 

FlickrStyle10K~\cite{gan2017stylenet} is collected and built on Flickr30K~\cite{hodosh2013framing} image caption dataset. The original FlickrStyle10K dataset has 10,000 pairs of images and stylized captions including humorous and romantic styles. However, only 7,000 pairs from ofﬁcial training set is now publicly accessible. We take the same processing method following Guo et al.\cite{guo2019mscap} and Zhao et al.\cite{Zhao2020MemCapMS} that randomly sample 6,000 images as our insufficient training data and the rest images are used for testing. To obtain the optimal setting of the hyper-parameters, we randomly split 10\% samples of the training set as the validation set. 

SentiCap~\cite{mathews2016senticap} constructs an additional set of captions with positive and negative sentiments using images from the MSCOCO~\cite{lin2014coco} validation partition. The positive and negative subsets contain 998/673 and 997/503 images for training/testing, respectively. We randomly sample 100 images from each of the training splits as the validation set.

In all experiments, we construct the large-scale paired factual corpus with the MSCOCO training partition and Flickr30K dataset excluding images and corresponding factual captions in the above stylized image caption datasets.

\subsection{Experimental Models}

To validate the effectiveness and applicability of our method, we implement the proposed data augmentation framework on following representative models, which can be classified into the three categories: supervised models (NIC and Up-Down), unsupervised models (StyleNet), and Transformer-based models (SAN and VinVL).

\begin{itemize}
    \item \emph{Neural Image Caption (NIC)}~\cite{Vinyals2015nic} is a standard encoder-decoder image caption pipeline, which encodes the image using CNN and decodes it using LSTM.
    \item \emph{Up-Down}~\cite{Anderson2018butd} employs a bottom-up and top-down attention mechanism with the two-layer LSTM, which enables attention to be calculated more naturally at the level of objects and other salient regions.
    \item \emph{StyleNet}~\cite{gan2017stylenet} devises a model component, named factored LSTM, which automatically distills the style factors in the monolingual text corpus. This model can explicitly control the style in the caption generation process so as to produce attractive visual captions with the desired style.
    \item \emph{Self-Attention Network (SAN)}~\cite{guo2020san} consists of an image encoder and a caption decoder, both of which are composed of a stack of transformer blocks. The inputs to the encoder are the region-based  visual features and the decoder takes the attended visual features and the embeddings of the previous words to predict the next word recursively. 
    \item \emph{VinVL}~\cite{zhang2020vinvl} is a powerful multi-modal pre-trained model for visual-language tasks, which is pre-trained on a large-scale text-image corpora and achieves new state-of-the-art results on several public benchmarks include image captioning.
\end{itemize}

\subsection{Comparison Models}
We also compare our approach with several previous state-of-the-art methods for stylized image captioning.
\begin{itemize}

\item Style-Factual LSTM (SF-LSTM)~\cite{chen2018factual} proposes an adaptive learning approach based on a reference factual model and provides factual knowledge to the model as the model learns from stylized caption labels, and can adaptively compute how much information to supply at each time step.

\item MSCap~\cite{guo2019mscap} proposes an adversarial learning framework with several well-designed modules to generate visually-grounded and style-controllable captions, which is trained on unpaired stylized corpus.

\item MemCap~\cite{Zhao2020MemCapMS} is a novel stylized image captioning method that explicitly encodes the knowledge about linguistic styles with memory mechanism, which is also trained on unpaired stylized corpus.

\end{itemize}

\subsection{Evaluation Metrics}
We evaluate our method on several baseline models from two aspects: the ability of generating fluent and accurate captions, and the performance of generating captions in target linguistic style. Following the related baseline work~\cite{guo2019mscap, Zhao2020MemCapMS}, we measure the sentence fluency and relevancy using the metrics of BLEU~\cite{DBLP:conf/acl/PapineniRWZ02} (including BLEU-1 and BLEU-3), METEOR~\cite{DBLP:conf/acl/BanerjeeL05}, CIDEr~\cite{vedantam2015cider}.

In addition, we also measure the sentence stylishness using the style classification accuracy (cls.) and the average perplexity (ppl.) following~\cite{guo2019mscap} and~\cite{Zhao2020MemCapMS}. Accuracy of style classification is defined as the percentage of sentences that correctly reflect the desired styles. For that purpose, we use BERT to train a style classifier for each of four styles with stylized sentences from FlickrStyle10K and SentiCap datasets and factual sentences from MSCOCO dataset. These classifiers achieve accuracies of 96.5\%, 97.4\%, 99.3\%, and 98.8\% on the test sets of humorous, romantic, negative, and positive styles respectively. As for the average perplexity, we use the SRILM toolkit to train a tri-gram based statistical language model for each of four styles on corresponding stylized sentences. The language models achieve perplexities of 76, 61, 47, and 52 on the test sets of four styles respectively. A lower score indicates that the generated sentences are more fluent and better reflect the desired linguistic style. 

\section{Results and Analysis}
\subsection{Quantitative Analysis}
\textbf{Main Results.} We report the results of several experimental models trained with the augmented and original data in Table \ref{tab:res-main}. As expected, all models trained with the augmented data substantially outperform those with the original data. Even for the traditional NIC model, both sentence stylishness (measured by cls.) and sentence relevancy (measured by BLEU, CIDEr, and METEOR) are greatly improved compared to the model trained on the original data. For example, under the evaluation of cls. and CIDEr, NIC achieves 6\% improvement and 14.7\% improvement on humorous style. Benefiting from the transformer architecture and pre-training technology, SAN and VinVL achieve nearly the best results. For example, VinVL achieves the best scores of all metrics on negative style compared with other models except for perplexity (ppl.). As for the perplexity metric, we claim that it does not correlate entirely well with stylishness, since the average perplexities on the test set of four styles are much higher than the results of Table \ref{tab:res-main}.

\begin{table}[]
\centering
\caption{Performance comparison with state-of-the-art methods. ``Ours(SAN)'' uses the augmented data for training.}
\label{tab:res-sota}
\resizebox{\linewidth}{!}{%
\begin{tabular}{llcccccc}
\hline
Style                  & Model   & B-1  & B-3  & M    & C    & cls. & ppl.          \\ \hline
\multirow{4}{*}{Humor} & SF-LSTM & 27.4 & 8.5  & 11.0 & 39.5 & -    & -             \\
                       & MSCap   & 16.3 & 1.9  & 5.3  & 15.2 & 91.3 & 22.7          \\
                       & MemCap  & 19.9 & 4.3  & 7.4  & 19.4 & 98.9 & 16.4          \\
 & \textbf{Ours~(SAN)} & \textbf{29.5} & \textbf{9.9}  & \textbf{12.5} & \textbf{47.2} & \textbf{99.4}  & \textbf{13.7} \\ \hline
\multirow{4}{*}{Roman} & SF-LSTM & 27.8 & 8.2  & 11.2 & 37.5 & -    & -             \\
                       & MSCap   & 17.0 & 2.0  & 5.4  & 10.1 & 88.7 & 20.4          \\
                       & MemCap  & 21.2 & 4.8  & 8.4  & 22.4 & 98.7 & 14.4          \\
 & \textbf{Ours~(SAN)} & \textbf{30.9} & \textbf{10.9} & \textbf{13.0} & \textbf{53.3} & \textbf{99.6}  & \textbf{13.1} \\ \hline
\multirow{4}{*}{Pos}   & SF-LSTM & 50.5 & 19.1 & 16.6 & 60.0 & -    & -             \\
                       & MSCap   & 46.9 & 16.2 & 16.8 & 55.3 & 92.5 & 19.6          \\
                       & MemCap  & 50.8 & 17.1 & 16.6 & 54.4 & 99.8 & 13.0          \\
 & \textbf{Ours~(SAN)} & \textbf{53.0} & \textbf{23.4} & \textbf{18.1} & \textbf{72.0} & \textbf{100.0} & \textbf{11.7} \\ \hline
\multirow{4}{*}{Neg}   & SF-LSTM & 50.3 & 20.1 & 16.2 & 59.7 & -    & -             \\
                       & MSCap   & 45.5 & 15.4 & 16.2 & 51.6 & 93.4 & 19.2          \\
                       & MemCap  & 48.7 & 19.6 & 15.8 & 60.6 & 93.1 & \textbf{14.6} \\
 & \textbf{Ours~(SAN)} & \textbf{51.2} & \textbf{20.5} & \textbf{17.6} & \textbf{67.0} & \textbf{100.0} & 14.8          \\ \hline
\end{tabular}%
}
\end{table}

\begin{figure*}
    \centering
    \includegraphics[width=1\textwidth]{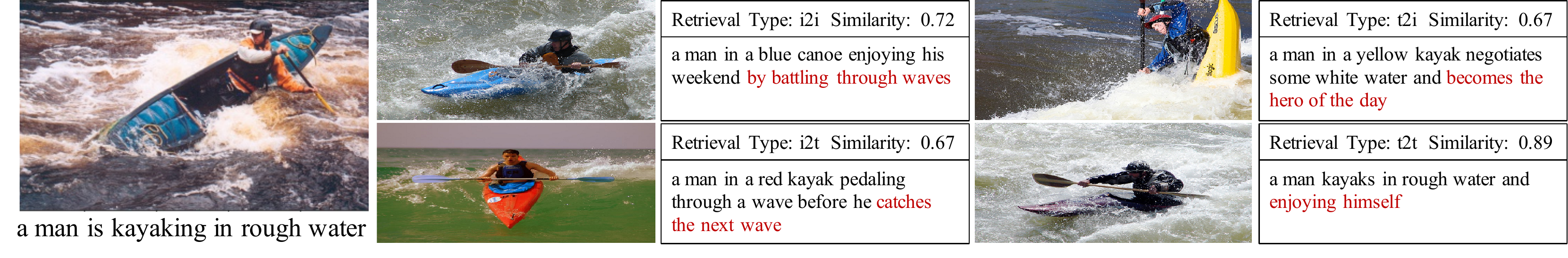}
    \caption{Retrieved stylized image-caption pairs with four types of retrieval. The left side is the queried image-caption factual pair. Words in red represent the style phrase. The similarity means the cosine similarity between the query and the result.}
    \label{fig:case-retrieval}
\end{figure*}

\begin{figure*}
    \centering
    \includegraphics[width=\linewidth]{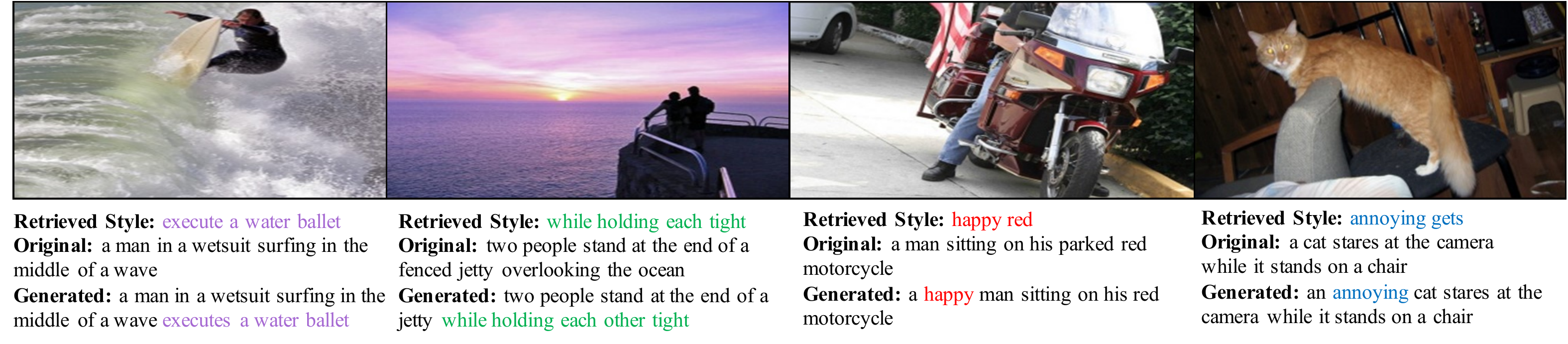}
    \caption{Augmentation examples generated by our framework. Words in purple, green, red, and blue represent the retrieved humorous, romantic, positive, and negative style phrases, respectively. Although there are some noise in the retrieved style parts, our generator can refine them correctly, e.g., \textit{each tight} $\rightarrow$ \textit{each other tight}.}
    \label{fig:case-example}
\end{figure*}

However, a seemingly abnormal observation is that StyleNet trained on the augmented data leads to a performance drop with respect to sentence relevancy but a great performance increase on sentence stylishness. We attribute it to its special training process, i.e., cross-training on a large number of factual captions and a small number of style captions. The imbalance of data leads to an excessive bias towards factual information and neglect of style expression. With augmented data, it can better balance the two kinds of information, although at the expense of some performance related to sentence relevance. We also observe that VinVL trained on the original data has achieved 96.6 on the evaluation of cls. for the romantic style. There is only 0.2\% improvement with augmented data. Combined with the result of CIDEr, it is obvious that VinVL trained on the original romantic data prefers to improve sentence stylishness over sentence relevance. 

From the above observations, we conclude that the key to this task is how to make a good trade-off between sentence relevance and sentence stylishness. It is empirically confirmed that our augmentation approach can generate more stylized samples for doing a better trade-off between the above aspects and training a more powerful model.

\textbf{Comparison with State-of-the-Arts.}
We select SAN with augmented data to compare with several state-of-the-art methods, i.e., SF-LSTM, MSCap, and MemCap. Table \ref{tab:res-sota} shows that SAN outperforms previous state-of-the-art models by a significantly large margin. In addition, regardless of the effect of the transformer architecture, NIC trained with augmented data also outperforms them across most metrics, as shown in Table \ref{tab:res-main}. The above results validate the effectiveness of our data augmentation framework. Meanwhile, those previous models can also be trained on our augmented data to obtain greater improvements without any architecture changes.

Compare MemCap/MSCap with baselines trained with original data, another seemingly abnormal observation is that there is a huge gap in sentence relevance. We claim that these well-designed models tend to sacrifice certain sentence relevance and improve the sentence stylishness. Baseline models trained simply with insufficient parallel data have difficulties in capturing stylishness and tend to improve relevance. Our ERG data augmentation framework is able to make a good trade-off between two aspects.

\begin{table}[]
\centering
\caption{Visualizations of the attention weights of stylized sentences produced by the proposed extractor.}
\resizebox{1\linewidth}{!}{
\begin{tabular}{ll}
\hline
Style & Stylized Sentences  \\ \hline
Humor & {\setlength{\fboxsep}{0pt}\colorbox{white!0}{\parbox{0.4\textwidth}{
\colorbox{purple!3.229363203048706}{\strut a} \colorbox{purple!0.8899415135383606}{\strut boy} \colorbox{purple!0.27412575483322144}{\strut wearing} \colorbox{purple!0.5439942479133606}{\strut an} \colorbox{purple!1.7077363729476929}{\strut orange} \colorbox{purple!0.7479191422462463}{\strut coat} \colorbox{purple!7.895837783813477}{\strut winks} \colorbox{purple!2.493382453918457}{\strut at} \colorbox{purple!3.6198627948760986}{\strut the} \colorbox{purple!6.341608047485352}{\strut camera} \colorbox{purple!14.312434196472168}{\strut suspiciously} 
}}}                            \\ 
Roman & {\setlength{\fboxsep}{0pt}\colorbox{white!0}{\parbox{0.4\textwidth}{
\colorbox{green!1.0927637815475464}{\strut a} \colorbox{green!0.4238969683647156}{\strut young} \colorbox{green!0.1529526561498642}{\strut boy} \colorbox{green!0.07608707994222641}{\strut in} \colorbox{green!0.1326397806406021}{\strut a} \colorbox{green!0.0435938723385334}{\strut red} \colorbox{green!0.057808808982372284}{\strut shirt} \colorbox{green!0.3649689853191376}{\strut plays} \colorbox{green!0.06060484051704407}{\strut on} \colorbox{green!0.07827157527208328}{\strut a} \colorbox{green!0.10246968269348145}{\strut mini} \colorbox{green!0.03295515105128288}{\strut trampoline} \colorbox{green!0.07896443456411362}{\strut in} \colorbox{green!0.036202095448970795}{\strut a} \colorbox{green!0.27582064270973206}{\strut grassy} \colorbox{green!0.5329645872116089}{\strut field} \colorbox{green!8.5827687978744507}{\strut having} \colorbox{green!15.258026123046875}{\strut fun} \colorbox{green!11.512611389160156}{\strut with} \colorbox{green!19.00178337097168}{\strut some} \colorbox{green!18.742427825927734}{\strut friends} 
}}}                             \\ 
Pos   & {\setlength{\fboxsep}{0pt}\colorbox{white!0}{\parbox{0.7\linewidth}{
\colorbox{red!1.6779893636703491}{\strut a} \colorbox{red!38.8120231628418}{\strut pretty} \colorbox{red!0.5476325154304504}{\strut woman} \colorbox{red!0.8662737607955933}{\strut smiling} \colorbox{red!0.24718347191810608}{\strut on} \colorbox{red!0.2387496829032898}{\strut her} \colorbox{red!31.872472763061523}{\strut favorite} \colorbox{red!0.6762199997901917}{\strut street} 
}}}                           \\ 
Neg   & {\setlength{\fboxsep}{0pt}\colorbox{white!0}{\parbox{0.7\linewidth}{
\colorbox{blue!0.5576068758964539}{\strut a} \colorbox{blue!26.3452091217041}{\strut bad} \colorbox{blue!1.575558066368103}{\strut image} \colorbox{blue!0.44520553946495056}{\strut of} \colorbox{blue!0.4287084937095642}{\strut a} \colorbox{blue!0.41090506315231323}{\strut person} \colorbox{blue!0.1240176260471344}{\strut in} \colorbox{blue!0.2184453159570694}{\strut a} \colorbox{blue!16.049272537231445}{\strut poor} \colorbox{blue!0.739860475063324}{\strut city} \colorbox{blue!0.1270936280488968}{\strut holding} \colorbox{blue!0.3164060115814209}{\strut out} \colorbox{blue!0.23832784593105316}{\strut stop} \colorbox{blue!0.24180223047733307}{\strut sign} \colorbox{blue!0.16791932284832}{\strut near} \colorbox{blue!0.28171995282173157}{\strut a} \colorbox{blue!2.0612316131591797}{\strut child} 
}}}                         \\ \hline
\end{tabular}
}
\label{tab:case-attention}
\end{table}

\subsection{Qualitative Analysis}
We visualize the heatmap of words from the stylized sentences and their attention weights from the extractor corresponding to $[CLS]$ token, as illustrated in Table~\ref{tab:case-attention}. According to the heatmap, whether each word belongs to the style phrase can be seen intuitively. It is obvious that attention weights of sentences highly correlate with sentence stylishness. It confirms that our extractor is able to effectively discern the style phrase of the sentence.

As shown in Figure~\ref{fig:case-retrieval}, we also presents a typical retrieval result of our multi-modal scene retriever. From the retrieval results, we can see our retriever can find more stylized matching pairs with low noise. In this way, we can add more diverse styles to the factual captions in the large-scale factual corpus.

Figure~\ref{fig:case-example} presents some augmented examples to illustrate the effectiveness of our augmentation approach intuitively.
Our framework can generate grammatically correct and diversified captions for scenes in the large factual dataset, even if style phrases extracted and retrieved from the original stylized data are slightly noisy.

\begin{table}[]
\centering
\caption{Effect of Different Retrieval Types.}
\label{tab:ablation-retrieval}
\begin{tabular}{l|cllcc}
\hline
Retrieval Type    & \#Samples      & \multicolumn{1}{c}{B-3} & \multicolumn{1}{c}{M} & C              & cls.          \\ \hline
image-to-image~(i2i) & 7087  & 9.55 & 11.87 & 46.03 & 94.1 \\
image-to-text~(i2t)  & 3955  & 9.36 & 11.76 & 44.88 & 91.4 \\
text-to-image~(t2i)  & 1382  & 8.96 & 11.49 & 44.73 & 89.7 \\
text-to-text~(t2t)   & 29116 & 9.73 & 12.18 & 46.73 & 96.9 \\ \hline
\textbf{Ours~(all)} & \textbf{37699} & \textbf{9.89}           & \textbf{12.44}        & \textbf{47.17} & \textbf{97.5} \\ \hline
\end{tabular}
\end{table}

\subsection{Ablation Studies}

We further conduct experiments to validate the effectiveness of different components. Unless otherwise specified, all the
experiments are carried out using the simplest NIC model with augmented data on romantic style.

\begin{table}[]
\centering
\caption{Effect of Different Caption Generators.}
\begin{tabular}{l|lccccc}
\hline
Style                  & Generator   & B-3           & M              & C              & cls.          & ppl.          \\ \hline
\multirow{2}{*}{Humor} & T5~(SEQ2SEQ) & 9.25          & \textbf{11.86} & 43.09          & \textbf{98.9} & 9.9          \\
                       & GPT-2~(LM)   & \textbf{9.57} & 11.76          & \textbf{45.21} & 97.5          & \textbf{9.2} \\ \hline
\multirow{2}{*}{Roman} & T5~(SEQ2SEQ) & \textbf{9.93} & 12.38          & \textbf{49.01} & 96.4          & 8.2          \\
                       & GPT-2~(LM)   & 9.89          & \textbf{12.44} & 47.17          & \textbf{97.5} & \textbf{8.0} \\ \hline
\end{tabular}
\label{tab:ablation-generator}
\end{table}

\begin{figure*}	
	\centering
	\begin{subfigure}[t]{0.24\linewidth}
		\centering
		\includegraphics[width=\textwidth]{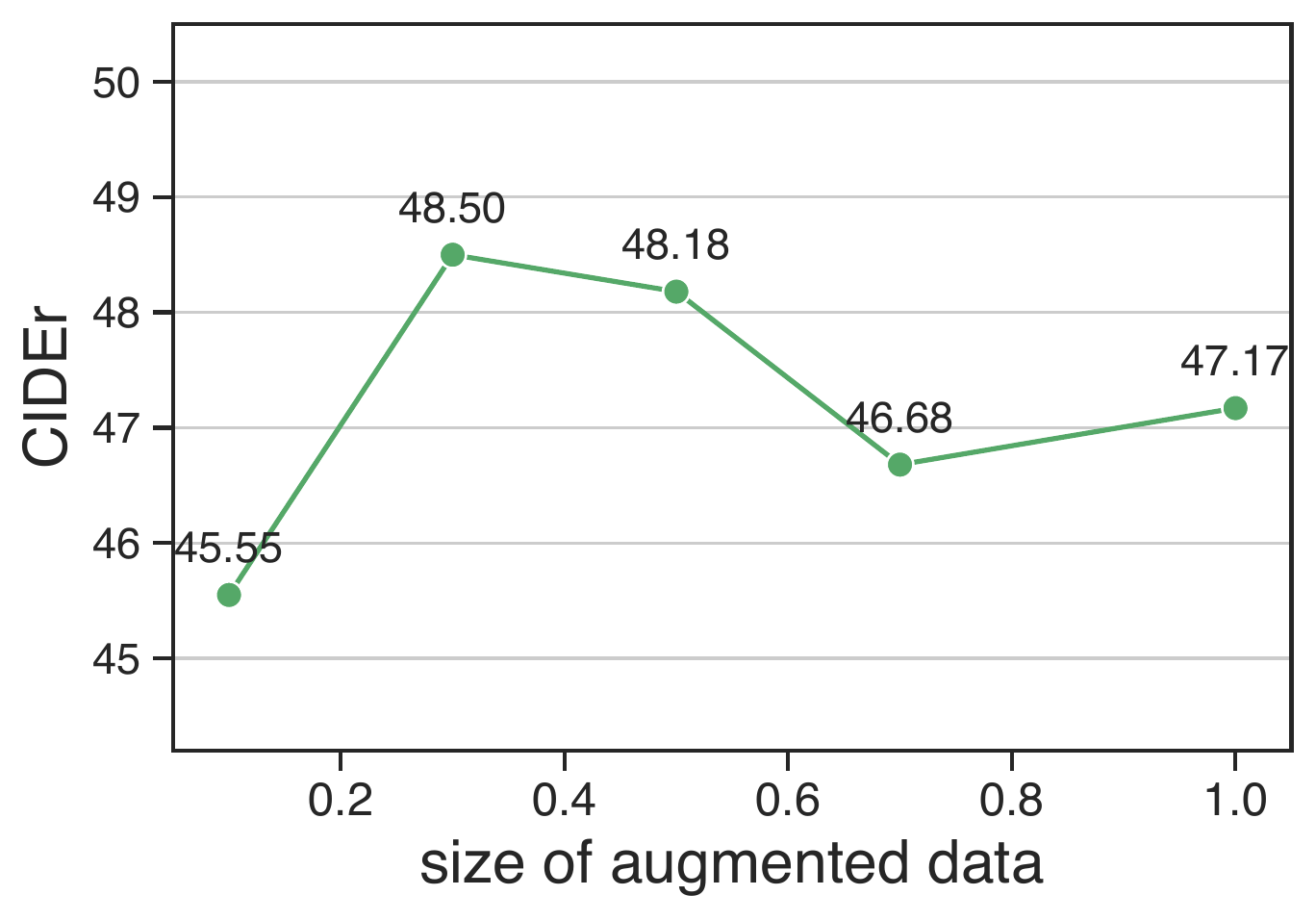}
		\caption{}\label{fig:ablation-size-cider}
	\end{subfigure}
	\begin{subfigure}[t]{0.24\linewidth}
		\centering
		\includegraphics[width=\textwidth]{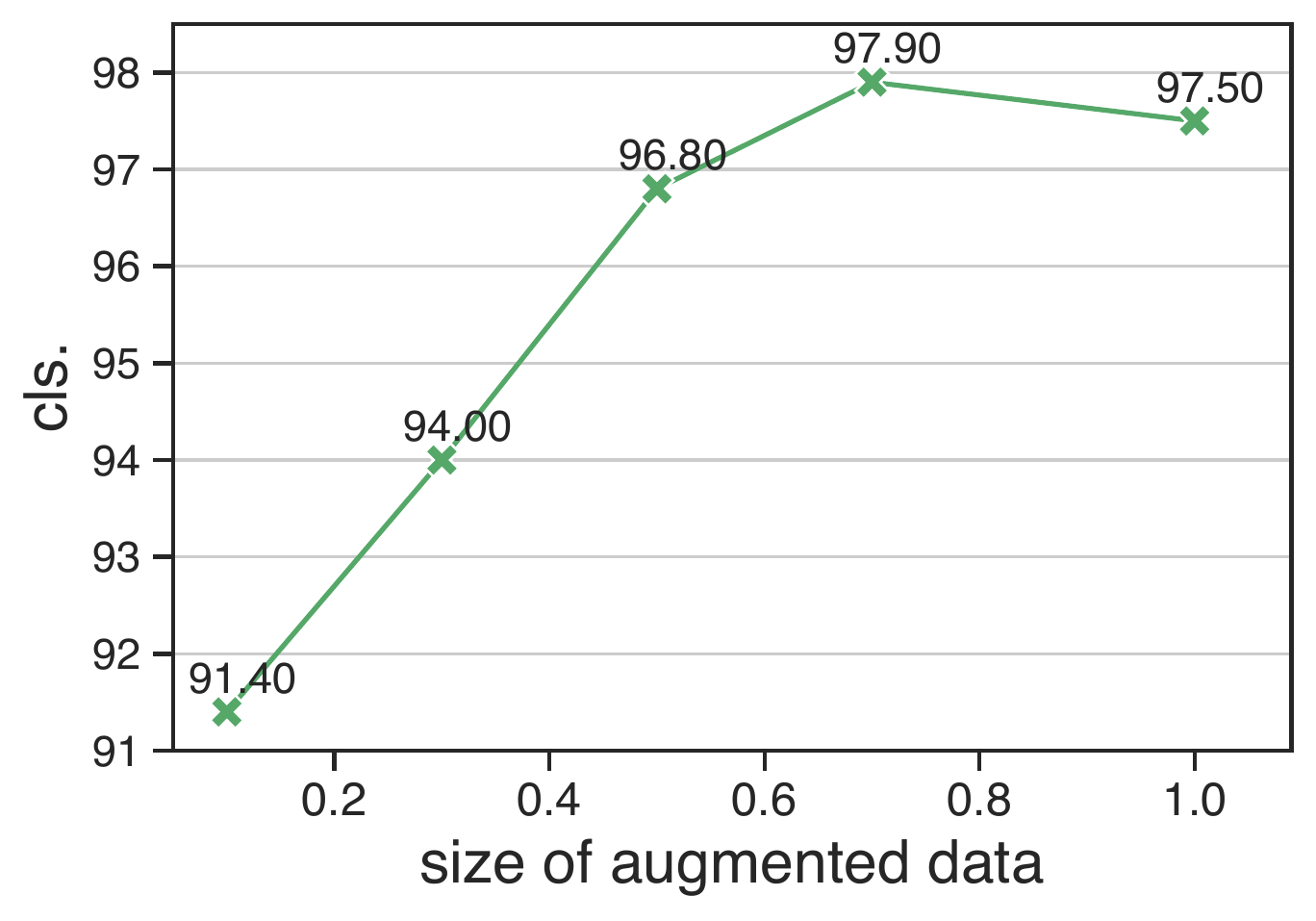}
		\caption{}\label{fig:ablation-size-cls}
	\end{subfigure}
	\begin{subfigure}[t]{0.24\linewidth}
		\centering
		\includegraphics[width=\textwidth]{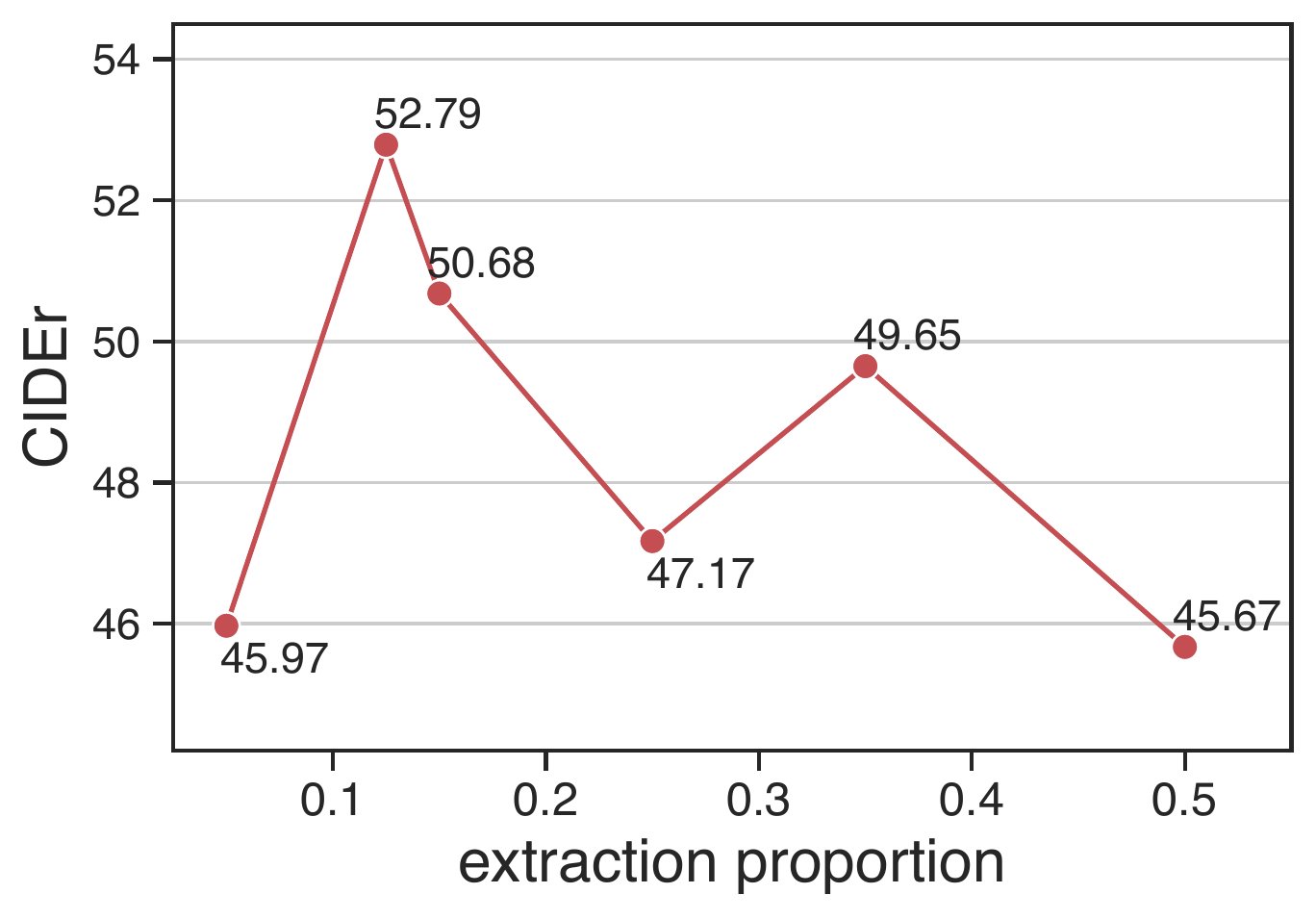}
		\caption{}\label{fig:ablation-split-cider}
	\end{subfigure}
	\begin{subfigure}[t]{0.24\linewidth}
		\centering
		\includegraphics[width=\textwidth]{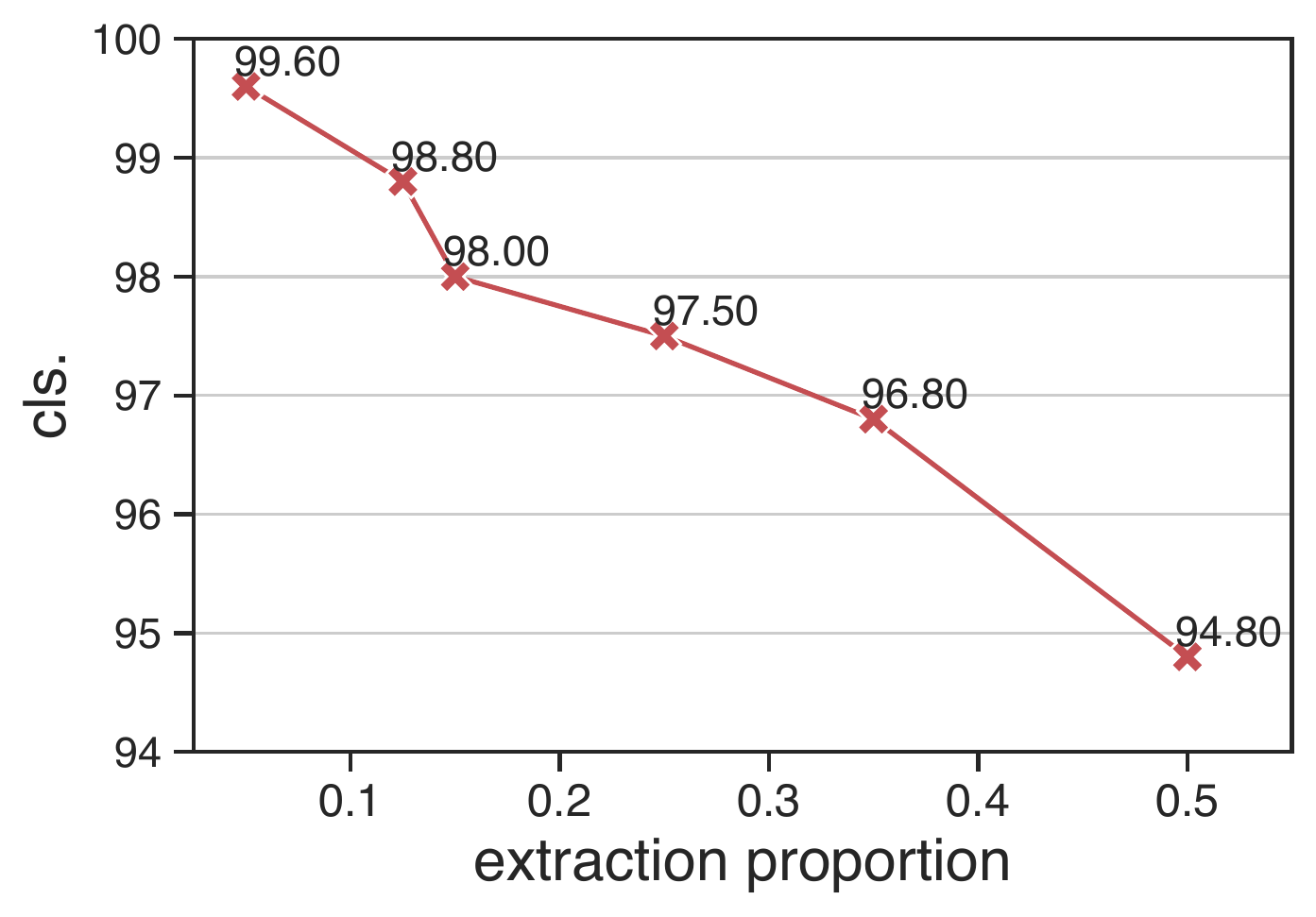}
		\caption{}\label{fig:ablation-split-cls}
	\end{subfigure}
	\caption{(a) CIDEr score with different size of augmented data. (b) cls. score with different size of augmented data. (c) CIDEr score with different extraction proportion. (d) cls. score with different extraction proportion.}
\end{figure*}

\textbf{Effect of the Size of Augmentation Data.} By training the NIC model on the 10\%, 30\%, 50\%, 70\%, and 100\% augmentation data, We investigate the impact of different-sized augmentation data. Specifically, the augmentation data are first sorted in ascending order of perplexity (see Sect. 4.4). Then, the top 10\%, 30\%, 50\%, 70\%, 100\% of sorted augmented data are selected to experiment. As shown in Figure~\ref{fig:ablation-size-cls}, we observe that the score of cls. increases as the size of augmentation data increases. When data augment to a certain size, the metric stops increasing. The phenomenon indicates the performance of this task can be boosted by increasing size of training data, but there is an upper bound of learning effective stylized knowledge. Meanwhile, there is a slight performance drop in terms of sentence relevance (from CIDEr of 48.50 to 47.17), since augmented data of larger size have more noise samples. 

\textbf{Effect of Extraction Proportion.} linguistic style is often localized to a small subset of words of a sentence. We investigate the impact of extraction proportion $\epsilon$ of the words ranked by the attention weights. Figure~\ref{fig:ablation-split-cls} shows that the score of the cls. metric gradually decreases as the proportion increases, which indicates that a larger proportion will introduce more noise. Meanwhile, Figure~\ref{fig:ablation-split-cider} shows that higher or lower proportions will degrade scores of sentence relevance to a certain extent. Nevertheless, NIC still achieves CIDEr of 45.97 and cls. of 94.80 with the proportion of 0.05 and 0.5, respectively, empirically confirming that our framework is insensitive to choice of the proportion within a reasonable range.

\textbf{Effect of Different Retrieval Types.} For validating the effectiveness of the multi-modal scene retriever, we also test the effects of different retrieval methods. Table \ref{tab:ablation-retrieval} reports performance scores of our approach with i2i, i2t, t2i and t2t retrieval respectively. From Table \ref{tab:ablation-retrieval}, we conclude that using multiple retrieval methods at the same time can get more augmented samples and improve the performance of the downstream task, in terms of cls. and CIDEr. From the retrieval cases in Figure~\ref{fig:case-retrieval}, it is more intuitively seen that our retriever can find more stylized matching pairs with low noise. 

\textbf{Effect of Different Caption Generators.} To present the flexibility of our augmentation framework, we analyze the impact of different state-of-the-art generative models (GPT-2 and T5) as the caption generator. While GPT-2 is a novel and powerful pre-trained language model, T5~\cite{Raffel2020t5} is another pre-trained model which adopts sequence-to-sequence architecture. As shown in Table~\ref{tab:ablation-generator}, the pre-trained models of two architectures have no obvious differences in downstream tasks, and both achieve good performance scores. From the above results, we conclude that our framework is powerful and flexible, which fully leverages the knowledge from large pre-trained transformers in a plug-and-play manner.

\section{Discussion}
In this section, we discuss the necessity of parallel data and the robustness of our framework, which are essential for validating the effectiveness and generalization of our proposed method.

\textbf{The unnecessity of parallel data.} The purpose of this paper is to augment the insufficient paired stylized data. However, if there are no stylized captions with aligned images at the beginning, our framework can also run well with unpaired stylized text and existing large-scale factual image-caption datasets. First, the process of extracting style phrases from the stylized text does not require aligned images. Our proposed emotional signal extractor just processes the stylized text to extract style phrases without requiring aligned images as the input. Second, when implementing the retriever with the text-to-text (t2t) or image-to-text (i2t) retrieval method, there are also no requirements for aligned images. From the results of Table~\ref{tab:ablation-retrieval}, we observe that only using text-to-text retrieval method can also achieve competitive results. Therefore, our framework can generate stylized paired data only with unpaired stylized text and existing large-scale factual image-caption pairs.

\textbf{The robustness of our framework.} Our framework consists of three separated modules running as an extract-retrieve-generate pipeline. Although the noise may be introduced in each process, our framework is still robust. From the augmented examples in Figure~\ref{fig:case-example}, our generator can produce grammatically correct sentences, even if there are some mistakes in the retrieved styles. For example, it discerns a missing word \textit{other} in ``\textit{while holding each tight}'', and modifies ``\textit{a cat}'' in the beginning of original sentence with ``\textit{an annoying cat}''. Meanwhile, the results of various experimental models in this task confirm that our framework is effective and robust.

\section{Conclusion}
\label{sec:conclusion}
In this paper, we propose a novel extract-retrieve-generate data augmentation framework for stylized image captioning. The proposed framework first extracts the style phrases from existing stylized sentences. Then the framework further retrieves a set of similar scenes through a plug-able multi-modal scene retriever. With the style phrases in the stylized captions about similar scenes, the framework generates corresponding stylized captions for images in the large-scale factual corpus.  Results of extensive experiments show that the framework can alleviate the data scarcity problem effectively. It also significantly boosts the performance of several existing image captioning models in both supervised and unsupervised settings, which outperforms the state-of-the-art stylized image captioning methods by a substantial margin.

\begin{acks}
We thank Tong Li, Yongliang Shen, Tian Lan, Chenyu Yan, Yan Yu, and Qinghao Ye for useful help. This work was supported by the NSFC projects (No. 62072399, No. U19B2042), Chinese Knowledge Center for Engineering Sciences and Technology, MoE Engineering Research Center of Digital Library, Alibaba-Zhejiang University Joint Research Institute of Frontier Technologies, and the Fundamental Research Funds for the Central Universities. 

\end{acks}
\newpage


\bibliographystyle{ACM-Reference-Format}
\balance
\bibliography{sample-base}

\appendix
\clearpage

\section*{Appendices} 

\subsection*{A. Augmentation Process}
Given an insufficient paired stylized data $\mathcal{D}_s$ and a large-scale paired factual data $\mathcal{P}$, we can obtain the augmented large-scale paired stylized data $\mathcal{D}_a$ through our Extract-Retrieve-Generate pipeline. Algorithm \ref{alg:augment} shows the whole augmentation process.

\begin{algorithm}
\caption{Augmentation Process}
\label{alg:augment}
\begin{flushleft}
    \textbf{Input:} \\
    \hspace*{\algorithmicindent} $\mathcal{D}_s$: an insufficient paired stylized image-caption data.\\
    \hspace*{\algorithmicindent} $\mathcal{P}$: a large-scale paired factual image-caption corpus.\\
    \hspace*{\algorithmicindent} $K$: the number of types of retrieval ($K=4$ means all scene retrievals in Table~\ref{tab:retrieval-type}). \\
    \hspace*{\algorithmicindent} $\epsilon$: the proportion of the most important words. \\
    \textbf{Output:}  \\
    \hspace*{\algorithmicindent} $\mathcal{D}_a$: a new large-scale paired stylized augmented data.
\end{flushleft}
\begin{algorithmic}[1]
\Procedure{Extract($\mathcal{D}_s$, $\epsilon$)}{}
    \State $\mathcal{D}'_s \gets \emptyset$
    \State \textbf{for} $(x^s, y^s)$ \textbf{in} $\mathcal{D}_s$
    \textbf{do}
    \State \hspace*{\algorithmicindent} $e^s \gets$ Extractor($y^s, \epsilon$)
    \State \hspace*{\algorithmicindent} $\mathcal{D}'_s \gets \mathcal{D}'_s \bigcup \{(x^s, y^s, e^s\}$
    \State \textbf{end}
    \State \textbf{return} $\mathcal{D}'_s$
\EndProcedure
\Procedure{Retrieve($\mathcal{D}'_s$, $\mathcal{P}, K$)}{}
    \State $\mathcal{P}' \gets \emptyset$
    \State \textbf{for} $(x^f, y^f)$ \textbf{in} $\mathcal{P}$
    \textbf{do}
    \State \hspace*{\algorithmicindent} $\{(x^s_i, y^s_i, e_i^s)|_{i=1}^{K}\} \gets $ Retriever($x^f, y^f$)
    \State \hspace*{\algorithmicindent} $\mathcal{P}' \gets \mathcal{P}' \bigcup \{(x^s_i, y^s_i, e_i^s, x^f, y^f)|_{i=1}^{K}\}$
    \State \textbf{end}
    \State \textbf{return} $\mathcal{P}'$
\EndProcedure
\Procedure{Generate($\mathcal{P}'$)}{}
    \State $\mathcal{D}_a \gets \emptyset$    
    \State \textbf{for} $(x^s, y^s, e^s, x^f, y^f)$ \textbf{in} $\mathcal{P}'$ \textbf{do}
    \State \hspace*{\algorithmicindent} $\hat{y}^s \gets $ Generator($y^f, e^s$)
    \State \hspace*{\algorithmicindent} $\mathcal{D}_a \gets \mathcal{D}_a \bigcup \{(x^f, \hat{y}^s)\}$
    \State \textbf{end}
    \State \textbf{return} $\mathcal{D}_a$
\EndProcedure

\State $\mathcal{D}'_s \gets$ Extract($\mathcal{D}_s$, $\epsilon$);
\State $\mathcal{P}' \gets$ Retrieve($\mathcal{D}'_s$, $\mathcal{P}, K$);
\State $\mathcal{D}_a \gets$ Generate($\mathcal{P}'$);
\State \textbf{return} $\mathcal{D}_a$;
\end{algorithmic}
\end{algorithm}

\subsection*{B. Implementation Details}
\subsubsection*{Data Augmentation}
For emotional signal extractor, we employ the pre-trained BERT-base model with the uncased version of tokenizer. We use a mini-batch size of 32 for romantic and humor styles and a mini-batch size of 64 for positive and negative styles. The learning rates of AdamW optimizers are set to 1e-5 and 2e-5, respectively. The numbers of fine-tuning epochs are both set to 3. The proportion of the most important words $\epsilon$ is set to 0.25. For multi-modal cross retriever, we use the pre-trained OSCAR as a unified multi-modal encoder to map texts and images into 768-dimensional vectors. For caption generator with emotional prompt, we employ the pre-trained GPT-2 model. The learning rate of AdamW optimizer is set to 5e-5 and the number of fine-tuning epochs is set to 1.

\subsubsection*{Experimental Models Settings}
For each image, we employ the Faster-RCNN~\cite{Ren2015rcnn} detector with ResNet-101 to extract the top $K = 36$ region proposals and obtain a 2048-dimensional feature for each region. In all experimental models except for SAN and VinVL, the hidden size of decoder is set to 512, which is also the size of word embeddings. As for the training process, we train these models under cross-entropy loss with a mini-batch size of 64, and Adam optimizer is used with a learning rate of 5e-4. We increase the probability of feeding back a sample of the word posterior by 0.05 every 5 epochs. In particular, for StyleNet, we take the same hyper-parameters and training methods suggested in~\cite{gan2017stylenet} to keep the original unsupervised settings. For SAN and VinVL, we also follow the hyper-parameters and settings suggested in~\cite{guo2020san} and~\cite{zhang2020vinvl}.

\end{document}